\documentclass{article}

     \usepackage[final]{neurips_2019}

\usepackage[utf8]{inputenc} 
\usepackage[T1]{fontenc}    
\usepackage{hyperref}       
\usepackage{url}            
\usepackage{booktabs}       
\usepackage{amsfonts}       
\usepackage{nicefrac}       
\usepackage{microtype}      

\usepackage{graphicx}
\usepackage{mathtools}
\usepackage{multirow}
\usepackage{float}
\usepackage{xcolor}

\title{LRS-DAG: Low Resource Supervised Domain Adaptation with Generalization Across Domains}

\author{%
  Rheeya Uppaal\thanks{Author webpage at: \url{http://uppaal.github.io/}} \\
  College of Information and Computer Sciences\\
  University of Massachusetts, Amherst\\
  Amherst, MA 01002\\
  \texttt{ruppaal@cs.umass.edu} \\
}

\begin{document}

\maketitle

\begin{abstract}
  Current state of the art methods in Domain Adaptation follow adversarial approaches, making training a challenge. Existing non-adversarial methods learn mappings between the source and target domains, to achieve reasonable performance. However, even these methods do not focus on a key aspect: maintaining performance on the source domain, even after optimizing over the target domain. Additionally, there exist very few methods in low resource supervised domain adaptation. This work proposes a method, LRS-DAG, that aims to solve these current issues in the field. By adding a set of "encoder layers" which map the target domain to the source, and can be removed when dealing directly with the source data, the model learns to perform optimally on both domains. LRS-DAG showcases its uniqueness by being a new algorithm for low resource domain adaptaion which maintains performance over the source domain, with a new metric for learning mappings between domains being introduced.
  We show that, in the case of FCNs, when transferring from MNIST to SVHN, LRS-DAG performs comparably to fine tuning, with the advantage of maintaining performance over the source domain. LRS-DAG outperforms fine tuning when transferring to a synthetic dataset similar to MNIST, which is a setting more representative of low resource supervised domain adaptation. 
\end{abstract}

\section{Introduction}

Domain adaptation (\citet{huang2007correcting}, \citet{ben2010theory}) aims to generalize a model from a source domain, with vast amounts of labelled data, to a target domain. Data in the target domain is almost always a large pool of unlabelled or partially labelled data. Domain Adaptation is typically achieved by learning a mapping between the domains. 

A popular way of learning these mappings is using Generative Adversarial Networks (\citet{goodfellow2014generative}), using the cycle consistency constraint from the CycleGAN (\cite{zhu2017unpaired}). This has shown promising results, (\cite{hoffman2017cycada}, \cite{liu2017unsupervised}); however, adversarial models are known to be notoriously hard to jointly train. (\cite{arjovsky2017towards})

There has been a series of non-adversarial approaches to learning domain mappings.
(\cite{hoshen2018nam}, \cite{long2015learning}, \cite{sun2016return}, \cite{sun2016deep}, \cite{haeusser2017associative}). However, all the aforementioned methods focus on the problem of large amounts of unlabelled data in the target domain. There exist many problems where collecting data at a large scale is hard. (\cite{motiian2017few}, \cite{patel2015visual}) There is limited work in this domain, (\cite{motiian2017few}, \cite{motiian2017unified}, \cite{hosseini2018augmented}), however, the typical approach is to use low capacity models to learn from this low resource data.

Additionally, there is almost no focus on maintaining performance on the source domain, while improving the target domain performance. This may be crucial in tasks where a unified model on both domains must be used, and thus a paradigm similar to multi-task learning would be required. (\cite{jiang2008literature}) For example, in the task of stellar classification, teaching the model to detect rare Supernovae should not deteriorate performance on detecting regular stars.

The method proposed in this work aims to address all of these problems: (1) Identifying  method for Supervised Domain Adaptation with limited labelled data, and (2) Creating a model that maintains performance on the source domain even after training on the target domain. In addition, the method also trains in a non-adversarial manner, which is an added advantage. The proposed method divides the network into two sets of layers, and a set of `Encoder' layers are inserted between the two sets of layers of the original network. The `Encoder layers' are trained to map the target distribution to the source (rather than mapping both into a domain invariant space, as with other methods), without changing the weights in the original network. Thus, simply removing the encoder layers assures the original optimal performance of the model on the source domain. The encoder layers are trained by minimizing a measure of distance between the two distributions: essentially, the Kullback–Leibler divergence, and second order statistics have been considered as objective functions. The proposed method has been implemented on two sets of source-target datasets, and two different neural network architectures. While the results are comparable to fine-tuning, the method maintains generalization across the domains, and shows promising results for future work. 

The main contributions of this work are: (1) Proposing and testing a set of new metrics for minimizing feature covariances across domains (2) Proposing a new method in the supervised low-resource domain adaptation setting, which in being non-adversarial is significantly easier to train (3) Proposing a model which maintains performance over the source domain when learning from the target, thus displaying better generalization across domains. It must also be noted that the proposed method can be made to handle the standard case of domain adaptation of high resource unlabelled data, with minor tweaks. The implementation of the unsupervised variant is part of future scope.

\section{Related Work}

Domain Adaptation primarily focuses on reducing a domain shift, in three major ways. The first approach applies a form of regularization to better fit the model to the target domain (\cite{aytar2011tabula}, \cite{bergamo2010exploiting}, \cite{becker2013non}). The second is to transform both domains into a domain invariant space, and make further inference for the specific task, based on the features in this space. A popular approach for this is to use Generative Adversarial Networks (\cite{goodfellow2014generative}), using the cycle consistency constraint from the CycleGAN (\cite{zhu2017unpaired}). This puts the constraint on a particular example, that is converted from the source to target and back to the source, such that the same example is obtained. (\cite{hoffman2017cycada}, \cite{liu2017unsupervised}) \cite{manders2018simple} align predicted class probabilities across domains to achieve state of the art results, in addition to being robust to overfitting. These class of methods consistently show state of the art results on standard benchmarks. However, all these methods train models adversarially with a minimax objective which makes reaching a optimum hard. (\cite{arjovsky2017towards}) In fact, recent work shows that the objective function of GANs have no optimum, and must be treated as equilibriation problems, showing that the use of traditional optimization algorithms on GANs is `broken'.
 (\cite{gemp2018global}, \cite{mescheder2017numerics} )

The third method is to find some form of a mapping from the source domain to the target domain. The proposed method roughly falls into this category, with the slight difference that a mapping from the target to the source domain is learnt. \cite{sun2016deep} and \cite{sun2017correlation} present methods closely related to the proposed method. They align the second order statistics of the source and target distributions with a non-linear transformation. The loss used is the CORAL loss, which is the Frobenious norm of the correlations of the source and target domains. \\ 
$$\mathcal{L}_{CORAL} = \frac{1}{4d^2} ||C_S - C_T||_F^2$$\\
 Unlike LRS-DAG, the method works on unsupervised domain adaptation. Additionally, the model does not learn maintain performance on the source domain while learning the target, and does thus not generalize across domains. It also uses a strong prior for both domains, by plugging in Alexnet at the base of the network.

\cite{haeusser2017associative} follow a very similar setting, with also using an unlabelled target domain. They learn statistically domain invariant  embeddings,  while  minimizing  the  classification error  on  the  labeled  source  domain. This models holds the same weaknesses as Deep CORAL.

\section{Methodology}

\begin{figure}
    \centering
    \includegraphics[scale=0.3]{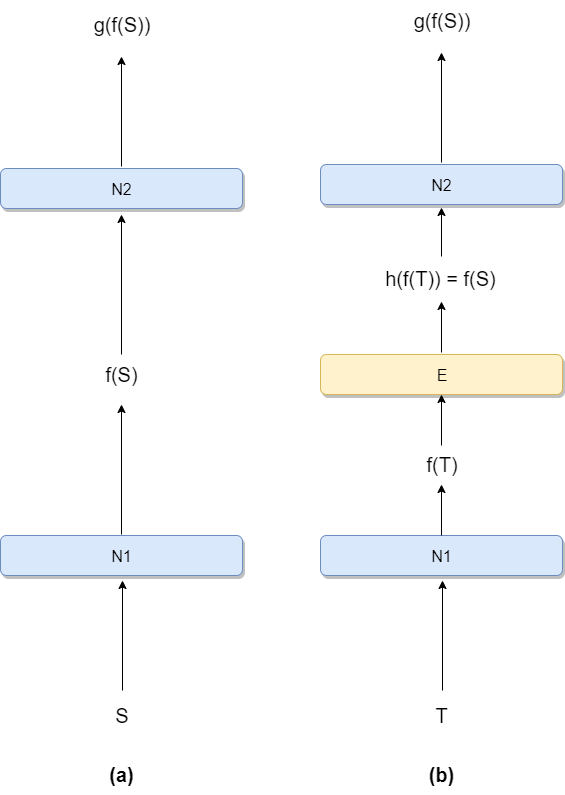}
    \caption{The architecture for the proposed methodology, with an arbitrary neural network trained on a classification task. The notation has been simplified such that S and T denote single datapoints from $\{S\}$ and $\{T\}$.}
\end{figure}

\subsection{LRS-DAG}


The LRS-DAG method works for an arbitrary model trained on the task of classification on the source data, $S$, where $s \in S$ and $s \in {\rm I\!R}^d$. The layers of the network are divided into two groups, $N1$ and $N2$. The model is trained in a standard manner, with the objective being to minimize an arbitrary classification loss. At the end of training, $N1$ learns a function $f: {\rm I\!R}^d \rightarrow {\rm I\!R}^k $, which maps $s$ to $f(s)$, $\forall s \in S$. Similarly, $N2$ learns the function $g: {\rm I\!R}^k \rightarrow {\rm I\!R}^c$, which maps $f(s)$ to $g(f(s)), \forall s$.

The key idea to generalizing across both domains is to keep the mappings created by $f$ and $g$ unaltered, and instead leverage them in their unaltered condition to optimize performance over the target data, $T$, where  $t \in T$ and $t \in {\rm I\!R}^d$. Thus, the weights of $N1$ and $N2$ are kept frozen in the next phase of training over the target domain data.

A new set of layers, the `Encoder layers', represented by $E$, are introduced between $N1$ and $N2$, in this phase (as shown in Figure 1). With the target domain, $E$ gets as input $f(t) \in {\rm I\!R}^k$, and must map that to $f(s)$. For this, $E$ is trained to learn a function $h: {\rm I\!R}^k \rightarrow {\rm I\!R}^k$ such that $h(f(t))=f(s)$. This would allow the input to $N2$ to be $f(s)$ regardless of the current domain, and $g$ can function in the same manner. The objective function to be minimized for training $E$ would thus be some measure of the difference between $f(s)$ and $h(f(t))$. Six measures of the objective function have been proposed:
\begin{itemize}
    \item CLS: $\mathcal{L} \coloneqq L_{class}$
    \item CLS+MSE: $\mathcal{L} \coloneqq \frac{1}{|T|}(\sum_{i=0}^{|T|} f(s_i) - h(f(t_i)))^2 + L_{class}$
    \item CLS+KL: $\mathcal{L} \coloneqq KL(f(S)||h(f(T))) + L_{class}$
    \item CLS+Norm: $\mathcal{L} \coloneqq \frac{1}{|T|}(\mu_{source} - \mu_{target})^2 + \frac{1}{|T|}(\sigma_{source} - \sigma_{target})^2 + L_{class}$
    \item CLS+KL-Rev: $\mathcal{L}$ := $KL(h(f(T))||f(S)) + L_{class}$ 
    \item CORAL: $\mathcal{L}$ := $L_{CORAL} + L_{class}$
\end{itemize}
where $L_{class}$ can be defined as any classification loss on $t \in T$, which in this case has been defined as the cross entropy loss between $\hat{y_{target}}$ and $t$, $L_{CORAL}$ is the CORAL loss from section 2 and $\mu_{source}$, $\sigma_{source}$, $\mu_{target}$ and $\sigma_{target}$ are the means and covariances of the source and target sets respectively. Methods CLS+KL and CLS+KL-Rev have both been included, since KL divergence is not symmetric. Method CORAL has been included as a comparison method. Since Deep CORAL is implemented in a different architecture and data setting than LRS-DAG, only the CORAL loss can be used as a comparison metric.

During inference, depending on the domain the model is currently being applied to, the encoder layers could be included or ignored from the forward pass ((a) and (b) in Figure 1). 

The proposed domain adaptation method is aimed to be model agnostic. Hence, it should perform for any arbitrary network, trained on the task of classification. For this reason, it has been tested with a basic fully connected network (for initial experimentation and proof of concept), and then with a standard CNN used for classification on the selected datasets.

\subsection{Other Aspects of the Training Regime}

A practical issue with the above loss metrics is that the number of examples in $S$ and $T$ vary, i.e. $|S| \neq |T|$, and the metrics use a one-to-one correspondence from the target domain. Two solutions were considered. The first is to parameterize the entire source distribution by finding $\mu_{source}$ and $\sigma_{source}, \forall s\in S$. Then sample $|T|$ points from a multivariate Gaussian with $(\mu_{source}, \sigma_{source})$. However, apart from the possibly incorrect assumption that the source data is Gaussian, these are just estimates of the mean and covariance of the true source distribution, and they may be biased. This might lead to $E$ learning a spurious function.

The other option is to simply sample $|T|$ points from $S$ (In practical implementation, however, we sample points for one minibatch at a time). However, an issue with this could be that a certain degree of information about the observed distribution would be lost. An extreme case of this would be where all the points are sampled from the tails of the observed source distribution, thus mapping $h$ to a distribution different from $f(S)$.

Both methods have been tested in the experimental section, and results are presented in Section 6. For simplicity, the first method shall be referred to as `indirect sampling' and the second method shall be referred to as `random sampling'.


\section{Datasets}

\paragraph{MNIST}The MNIST dataset contains 28x28 sized grayscale images of handwritten digits labelled from 0 to 9, and predefined training and testing splits of 60,000 and 10,000 examples apiece. The images were scaled to 32x32 and normalized. This has been used as the source domain.

\paragraph{SVHN}The Street View House Number dataset is a real-world image dataset obtained from Google Street View images. Like MNIST, it contains images of cropped digits between 0 and 9, but the images come from a significantly harder problem. The dataset consists of approximately 73,000 training images (out of which 10\% has been retained for the limited labelled data scenario) and 26,000 test images. The images were converted to greyscale and normalized. MNIST-SVHN is a standard benchmark for domain adaptation tasks, which is why these datasets have been used for initial testing.

\paragraph{Synthetic-MNIST}To see how LRS-DAG performs with different levels of domain shift, this dataset was created by applying a series of transformations on MNIST. Random horizontal flips over samples from the data were applied, and images were sheared. In addition to this, the brightness, contrast and saturation of images was randomly changed. Like with SVHN, only 10\% of the labelled training data was used.

Validation sets were made from the training splits for these datasets, and rolled back into the training sets after performing a grid search over the hyperparameter space, and judging performance over the validation set.

\begin{figure}[!htb]
    \minipage{0.2\textwidth}
      \includegraphics[scale=0.1]{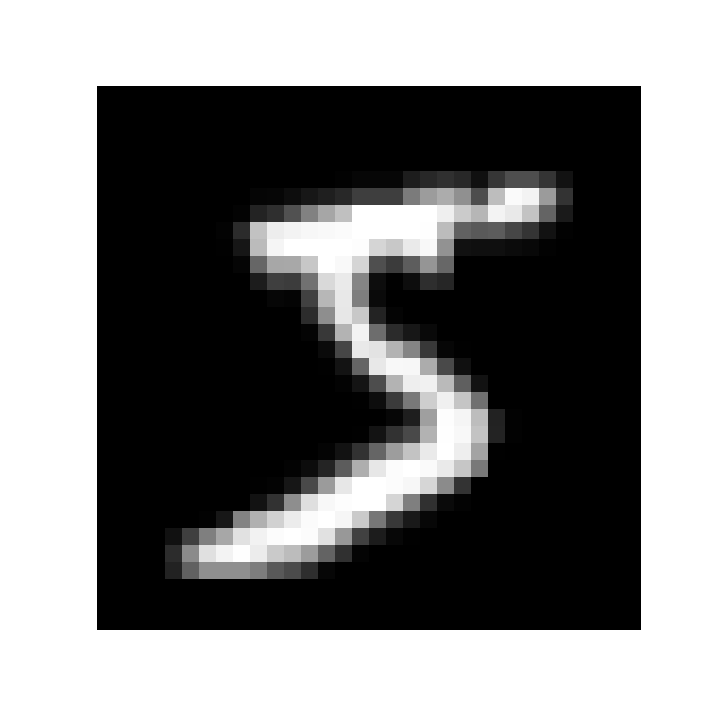} 
    \endminipage\hfill
    \minipage{0.2\textwidth}
      \includegraphics[scale=0.1]{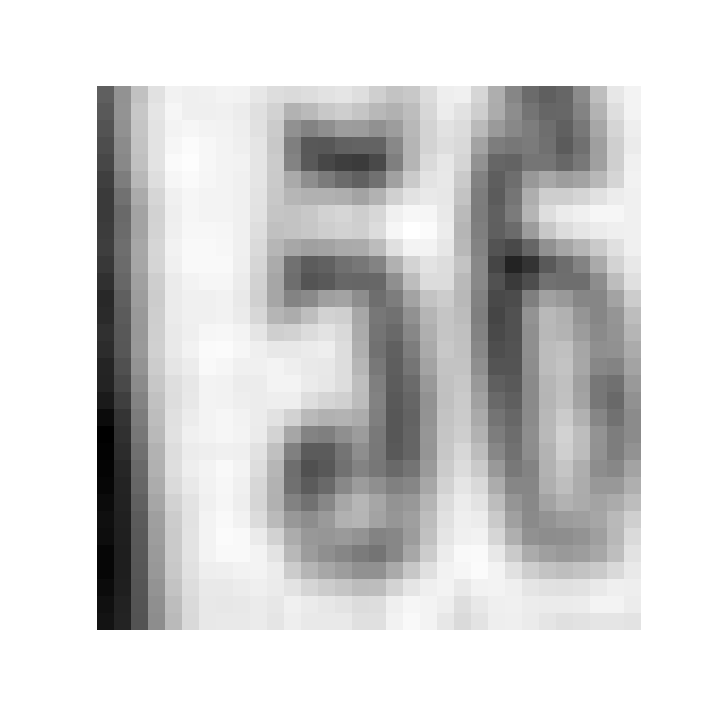}
    \endminipage\hfill
    \minipage{0.2\textwidth}%
      \includegraphics[scale=0.1]{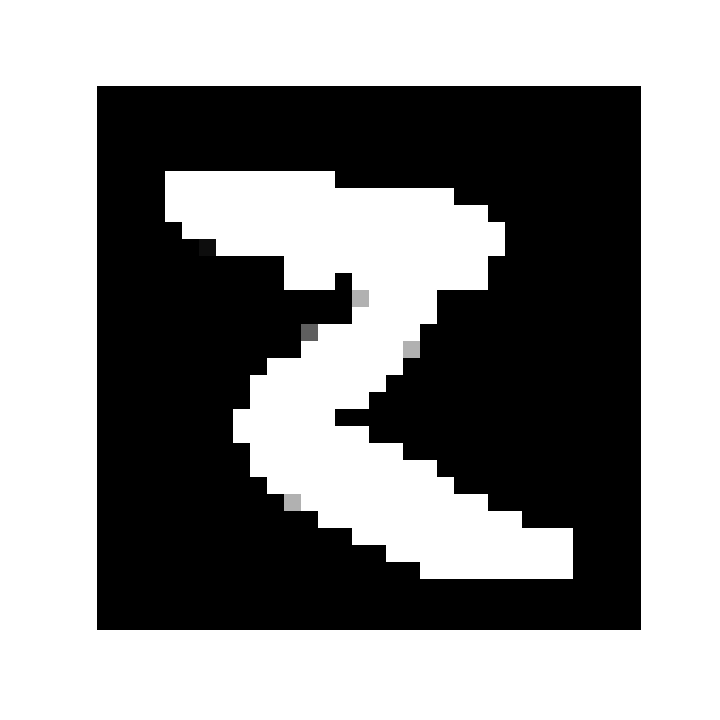} 
    \endminipage
    \minipage{0.2\textwidth}%
        \includegraphics[scale=0.2]{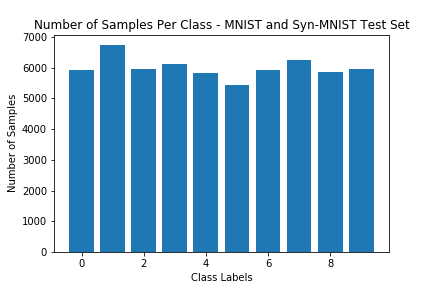}  
    \endminipage
    \minipage{0.2\textwidth}%
        \includegraphics[scale=0.7]{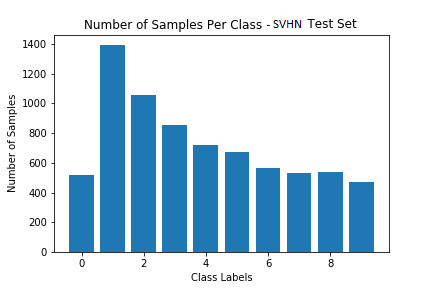} 
    \endminipage
        \caption{Left: Samples from the MNIST, SVHN and Syn-MNIST datasets. Right: Distributions of classes over the test data, for MNIST and SVHN.}
\end{figure}

\section{Experiments}

The goal of this series of experiments was to test the LRS-DAG method, with all its variants of loss functions. To show that the proposed training regime would work as an efficient form of Domain Adaptation, it has been tested over different models, and different sets of datasets. To test for the correctness of the hypothesis that the method is model agnostic, all experiments have been run for two networks:
\begin{itemize}
    \item \textbf{Model 1 - FCN}:  A fully connected network with 4 hidden layers. The output layer generates softmax predictions for all classes. $N1$, $E$ and $N2$ consist of the bottom two, middle two and top two layers of the network. $E$ is ignored when in the source domain. The network has no non-linearities.
    \item \textbf{Model 2 - CNN}: A CNN used for learning from domains of similar complexity. The network consists of 5 convolutional layers and one fully connected layer, each followed by ReLU non-linearity. Softmax is applied over the output of the last layer to give a confidence score for every class. As with FCN, $N1$, $E$ and $N2$ consist of the bottom two, middle two and top two layers of the network. $E$ takes and returns values of the same shape, in both models. 
\end{itemize}

The model is first trained on the source domain for 100 epochs. Following this, there are three main sets of experiments: $MNIST\rightarrow SVHN$ with FCN, $MNIST\rightarrow SVHN$ with CNN, $MNIST\rightarrow Syn-MNIST$ with CNN, where $\rightarrow$ signified transferring across domains. For each of these sets of experiments, all the loss functions (with the direct and indirect sampling methods described in Section 3.2) are tested. Additionally, they are compared with a series of baselines.

\paragraph{Basline Methods:} (1) \textbf{Source Trained}: The most rudimentary baseline method considered was training a model on the source dataset, and performing inference on the target with no additional training. This gives a lower bound on performance. (2) \textbf{Target Trained}: Train the model from scratch on the target domain. The high capacity model is likely to overfit to the data, thus performing poorly on the target test set. (3) \textbf{Finetune N2}: Finetune the weights of $N2$ on the target domain, after training the model on the source domain. This is akin to the most standard method of transfer learning, when limited labelled data is available. (4) \textbf{CORAL Loss}: Despite the Deep CORAL method (described in Section 2) being targeted towards the setting of a large pool of unlabelled data, the method is still most similar to LRS-DAG. For this reason, the CORAL loss has been fit into the LRS-DAG architecture as a loss function. This is expected to be the best performing method, as the loss minimizes second order statistics of the source and target distributions.

Additional points of note are that:
\begin{itemize}
    \item The method was implemented from scratch, using PyTorch 0.4, Scipy, Numpy and Scikit-Learn. No other existing implementations or frameworks were used.
    \item The accuracy of the model on the hidden test set of the target domain was used as a metric of the performance of a model. Confusion Matrices were also used to further analyse the methods, but have been excluded from this work for brevity.
    \item Hyperparameter tuning was done through a grid search over learning rate, weight decay, and the kind of optimizer. This performance was measured over the validation set, which was later rolled back into the training set for all methods. The validation set splits were stored and, for a particular dataset, the same data points were used as the validation set for all methods.
    \item The Adam optimizer was used for all models. On average, all methods for a particular model-source-target triplet required very similar hyperparameters.
    \item To account for the stochasticity arising from random weight initializations, every experiment has been run for three trials, and their averaged results have been showcased in Table 1.
    \item All models were trained until satisfying the stopping criteria of the difference in loss between two epochs being less than a particular threshold (thresholds varied, based on the type of loss function).
    \item Since intermediate features in a network are not probability distributions, and the method relies on the assumption they are distributions, and a softmax function is applied over the features after extracting them from the network, to convert them into a valid probability distribution.
\end{itemize}

\section{Results}


\paragraph{Experiment Set 1}: Indirect sampling of the source domain consistently outperforms random sampling. Hence, the information loss while sampling 10\% points from the source is relatively large. All proposed methods were expected to have similar outcomes, but the CLS+KL method ($KL(f(S)||(h(f(T))$) slightly outperforms the other methods. CLS+KL-Rev has a very similar performance since it is still a very similar notion of distance that is being minimized between both methods.

A point worth noting is that, while the  Target Trained baseline outperformed other methods, LRS-DAG with CLS+KL is almost the same as fine tuning. However, unlike fine tuning, the proposed method maintains generalization across both domains. 

Another notable point is the weak performance of CORAL based model in all three experiment sets. This may be because the Deep CORAL method simultaneously trains on the source and target domains, jointly minimizing estimators of the true $f(S)$, with roughly equal strength in both domains. With LRS-DAG, the estimate of $f(S)$ from the source is already very accurate, which might cause $h(f(S))$ to converge to an alternate value.

    \begin{table}[h]
        \centering
        \caption{Accuracies (in percentage, averaged over three trials) of baselines and proposed methods, for FCN, when transferring from MNIST to SVHN. Results of the two variants of the model: Without E (the model without the encoder layers, intended for inference over the source domain) and With E (the model with the encoder layers, intended for inference over the target domain) are shown.}
            \begin{tabular}{llllll}
             \toprule
             \multirow{2}{*}{Method} & \multirow{2}{*}{Sampling Method} & \multicolumn{2}{c}{Without $E$} & \multicolumn{2}{c}{With $E$}\\
             & & Source & Target & Source & Target \\ \midrule
             Target Trained & - & 25.44 & 13.92 & 6.39 & \textbf{34.52}\\ 
             Finetune N2 & - &  15.42 & 14.98 & 11.08 & 30.35  \\
             CLS & - & 91.91 & 13.93 & 11.46 & 29.97   \\ 
             CLS+MSE & Indirect & 91.91 & 13.93 & 13.53 & 29.64   \\ 
             CLS+MSE & Random & 91.91 & 13.93 &   11.31 & 29.38  \\ 
             CLS+KL & Indirect & 91.91 & 13.93 & 12.18 & \textbf{30.28}  \\ 
             CLS+KL & Random & 91.91 & 13.93 &  14.25 & 28.85 \\ 
             CLS+Norm & Indirect & 91.91 & 13.93 & 10.55 & 29.66  \\ 
             CLS+Norm & Random & 91.91 & 13.93 &  10.98 & 28.15 \\ 
             CLS+KL-Rev & Indirect & 91.91 & 13.93 & 11.75 & 29.79  \\ 
             CLS+KL-Rev & Random & 91.91 & 13.93 &  7.73 & 30.15 \\ 
             CORAL & Indirect & 91.91 & 13.93 & 11.35 & 19.59  \\ 
             CORAL & Random & 91.91 & 13.93 & 12.46 & 19.32 \\ 
             \bottomrule
        \end{tabular}
    \end{table}
    
    \begin{table}[h]
        \caption{Accuracies (in percentage, averaged over three trials) of baselines and proposed methods, for CNN, when transferring from MNIST to SVHN. Results of the two variants of the model: Without E (the model without the encoder layers, intended for inference over the source domain) and With E (the model with the encoder layers, intended for inference over the target domain) are shown.}
        \centering
        \begin{tabular}{llllll}
             \toprule
             \multirow{2}{*}{Method} & \multirow{2}{*}{Sampling Method} & \multicolumn{2}{c}{Without $E$} & \multicolumn{2}{c}{With $E$}\\
             & & Source & Target & Source & Target \\ \midrule
             Finetune N2 & - & 16.44 & 18.47 & 13.53 & \textbf{26.77}  \\ 
             CLS & - & 93.88 & 20.19 & 24.34 & 21.88  \\ 
             CLS+MSE & Indirect & 93.88 & 20.19 & 26.62 & 21.73   \\ 
             CLS+KL & Indirect & 93.88 & 20.19 & 28.52 & 21.85  \\ 
             CLS+Norm & Indirect & 93.88 & 20.19 & 60.79 & 21.29  \\ 
             CLS+KL-Rev & Indirect & 93.88 & 20.19 & 29.24 & \textbf{21.92}  \\ 
             CORAL & Indirect & 93.88 & 20.19 & 11.46 & 20.59  \\ \bottomrule
        \end{tabular}
    \end{table}

    \begin{table}[h]
        \centering
        \caption{Accuracies (in percentage, averaged over three trials) of baselines and proposed methods, for FCN, when transferring from MNIST to Syn-MNIST. Results of the two variants of the model: Without E (the model without the encoder layers, intended for inference over the source domain) and With E (the model with the encoder layers, intended for inference over the target domain) are shown.}
        \begin{tabular}{llllll}
             \toprule
             \multirow{2}{*}{Method} & \multirow{2}{*}{Sampling Strategy} & \multicolumn{2}{c}{Without $E$} & \multicolumn{2}{c}{With $E$}\\
             & & Source & Target & Source & Target \\ \midrule
             Finetune N2 & - &  89.27 & 65.42 & 86.12 & \textbf{77.33}  \\ 
             CLS & -  & 91.91 & 63.13 & 84.52 & 77.09  \\ 
             CLS+MSE & Indirect & 91.91 & 63.13 & 84.98 & 78.09   \\
             CLS+KL & Indirect & 91.91 & 63.13 & 85.19 & \textbf{78.14}  \\
             CLS+Norm & Indirect & 91.91 & 63.13 & 85.26 & 77.95  \\
             CLS+KL-Rev & Indirect & 91.91 & 63.13 & 84.88 & 78.11  \\
             CORAL & Indirect & 91.91 & 63.13 & 76.07 & 67.94  \\
             \bottomrule
        \end{tabular}
    \end{table}
    
    \begin{figure}[h]
        \minipage{0.33\textwidth}
          \includegraphics[scale=0.3]{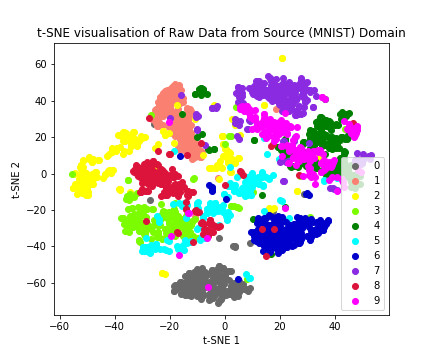} 
        \endminipage\hfill
        \minipage{0.33\textwidth}
          \includegraphics[scale=0.3]{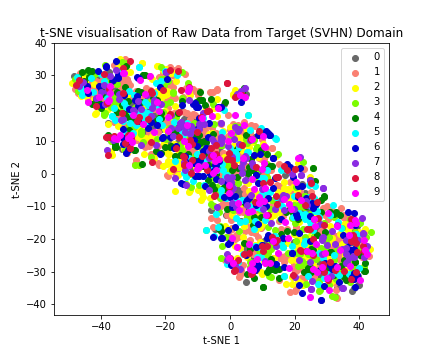}
        \endminipage\hfill
        \minipage{0.33\textwidth}%
          \includegraphics[scale=0.3]{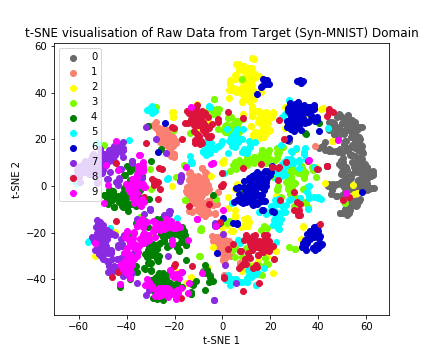} 
        \endminipage
        
        \minipage{0.33\textwidth}
          \includegraphics[scale=0.3]{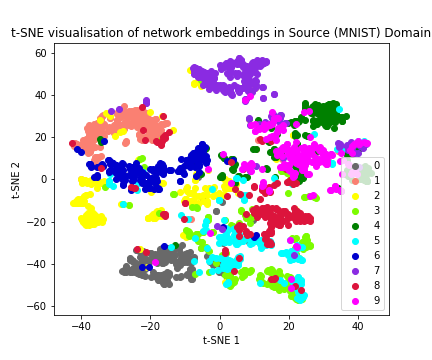} 
        \endminipage\hfill
        \minipage{0.33\textwidth}
          \includegraphics[scale=0.3]{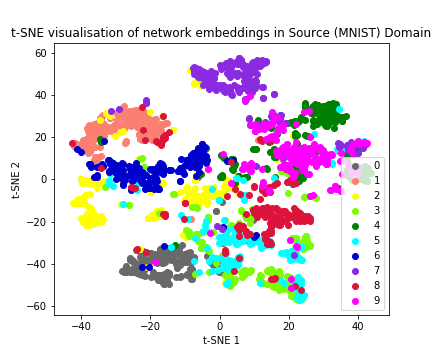}
        \endminipage\hfill
        \minipage{0.33\textwidth}%
          \includegraphics[scale=0.3]{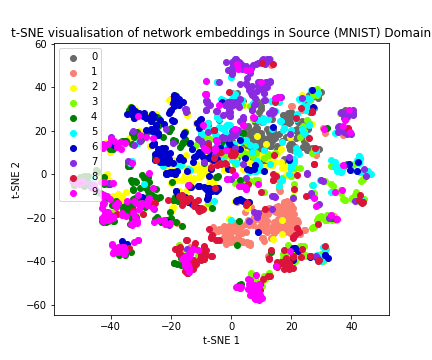} 
        \endminipage
        
        \minipage{0.33\textwidth}
          \includegraphics[scale=0.3]{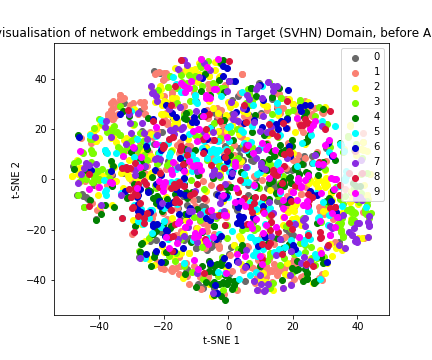} 
        \endminipage\hfill
        \minipage{0.33\textwidth}
          \includegraphics[scale=0.3]{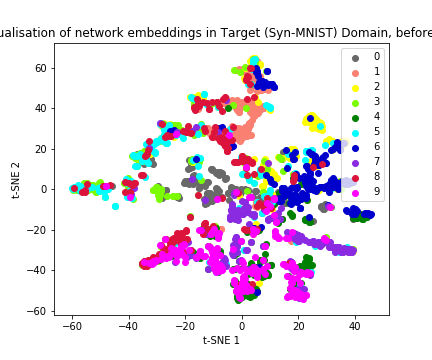}
        \endminipage\hfill
        \minipage{0.33\textwidth}%
          \includegraphics[scale=0.3]{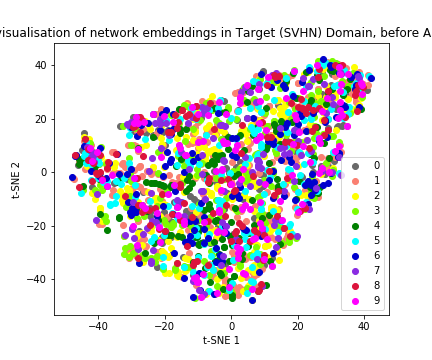} 
        \endminipage
        
        \minipage{0.33\textwidth}
          \includegraphics[scale=0.3]{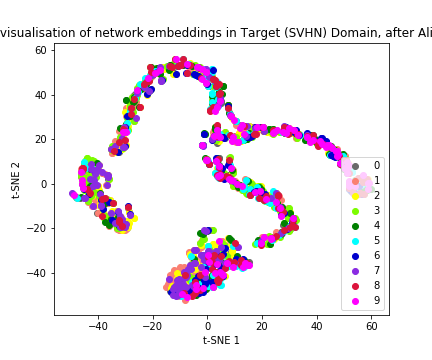} 
        \endminipage\hfill
        \minipage{0.33\textwidth}
          \includegraphics[scale=0.3]{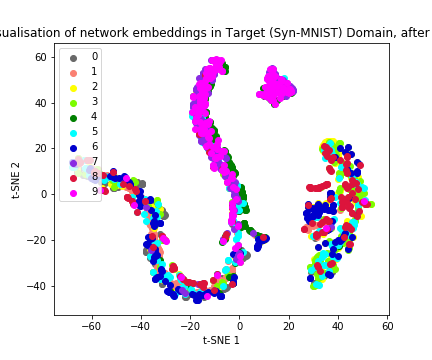}
        \endminipage\hfill
        \minipage{0.33\textwidth}%
          \includegraphics[scale=0.3]{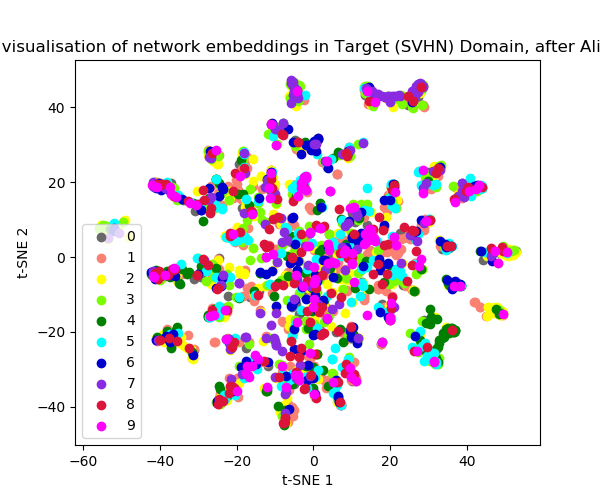} 
        \endminipage
    
        \caption{Top Row: t-SNE Visualizations of raw data from MNIST, SVHN and Syn-MNIST. Second row: $f(S)$ for source domain. Third row: $f(T)$ before training. Bottom: $h(f(T))$ after feature alignment. Left column: MNIST$\rightarrow$ SVHN with FCN. Middle column: MNIST$\rightarrow$ Syn-MNIST with FCN. Right column: MNIST$\rightarrow$ SVHN with CNN.}
        \label{tsne}
    \end{figure}

\paragraph{Experiment Set 2}: When using the CNN for adapting from MNIST to SVHN, it seems possible that the stopping criterion was not accurately applied. This would explain why the results on the target set in this experiment set is lower than the previous experiment set, despite the CNN being more powerful. Once again, the loss function based on KL divergence outperformed other proposed methods, and the CORAL loss. Additionally, the CLS+KL and CLS+KL-Rev methods have almost identical results here as well. The results of all methods are extremely similar in this set, leading to inconclusive results. However, finetuning clearly surpasses other methods. 

\paragraph{Experiment Set 3}: The Syn-MNIST domain has a lesser shift from MNIST than SVHN. Thus, the results seem more promising. Here, the CLS+KL and CLS+KL-Rev methods outperform finetuning, which is the most promising result so far. It is worth noting that this is a significant, as in low resource supervised domain adaptation, it is common to treat a highly similar domain as the source domain, and finetune over the target. LRS-DAG provides a clear benefit over other methods in this case.

\section{Discussion and Conclusion}


The LRS-DAG method seems comparable to fine-tuning, except for the case of the Syn-MNIST dataset, a closely related domain. However, the proposed method significantly outperforms CORAL, and most importantly, maintains generalization across both domains.

The method was inconclusive with CNNs. The performance may have been such due to a bad stopping criteria, or difficulty in aligning domains across convolutions. However, looking at the t-SNE plots of the aligned target domain after training (Figure \ref{tsne}, bottom right), the points of all classes have been clustered together. Thus, the proposed method of LRS-DAG shows promise. Taking the observations so far into account, and generating a better experimental setup, may provide more promising results in the future.

A point worth arguing would we whether it would make more sense to add $E$ to the top of the network, rather than making it handle intermediate features. A series of experiments (involving training different parts of the network and analyzing results) showed that the features across domains differ across lower levels, indicating the positioning of $E$ lower in the network in more beneficial (results excluded for brevity).

A possible path to pursue in the future would be to align $c$ different domains for each class, instead of an overall domain loss. Another field to explore would be mapping this model to the unsupervised domain adaptation setting. The method currently requires access to the source domain, when training on the target. Since access to the entire source domain is not always possible, another addition to the method could be to use the source domain embeddings to minimize domain difference, rather than using the data itself.

\clearpage



{\small
\bibliography{bibtex_lrs-dag}}

\end{document}